\documentclass[11pt]{article}

\usepackage[final]{acl}

\usepackage{times}
\usepackage{latexsym}

\usepackage[T1]{fontenc}

\usepackage[utf8]{inputenc}

\usepackage{microtype}

\usepackage{inconsolata}

\usepackage{graphicx}

\usepackage{subfig}
\usepackage{amsmath}
\usepackage{amsfonts}
\usepackage{cases}
\usepackage{booktabs}
\usepackage{multirow}
\usepackage{nicematrix}
\definecolor{grassgreen}{HTML}{E2F3DE}
\definecolor{gray}{HTML}{EFEFEF}

\title{GRASS: \underline{Gr}adient-based \underline{A}daptive Layer-wi\underline{s}e Importance \underline{S}ampling for Memory-efficient Large Language Model Fine-tuning}

\author{
 \textbf{Kaiyuan Tian\textsuperscript{1}},
 \textbf{Yu Tang\textsuperscript{2}},
 \textbf{Gongqingjian Jiang\textsuperscript{1}},
 \textbf{Baihui Liu\textsuperscript{1}},
\\
 \textbf{Yifu Gao\textsuperscript{1}},
 \textbf{Xialin Su\textsuperscript{1}},
 \textbf{Linbo Qiao\textsuperscript{1}\thanks{Corresponding authors.}},
 \textbf{Dongsheng Li\textsuperscript{1$\ast$}}
\\
 \textsuperscript{1}National University of Defense Technology\\
 \textsuperscript{2}Information Support Force Engineering University
\\
 \texttt{\{kyt,tangyu14,jianghydro,lbh,gaoyifu,suxialin25,qiao.linbo,dsli\}@nudt.edu.cn}
}

\begin{document}

\maketitle
\begin{abstract}
Full-parameter fine-tuning of large language models is constrained by substantial GPU memory requirements. 
Low-rank adaptation methods mitigate this challenge by updating only a subset of parameters. However, these approaches often limit model expressiveness and yield lower performance than full-parameter fine-tuning.
Layer-wise fine-tuning methods have emerged as an alternative, enabling memory-efficient training through static layer importance sampling strategies. However, these methods overlook variations in layer importance across tasks and training stages, resulting in suboptimal performance on downstream tasks.
To address these limitations, we propose GRASS, a gradient-based adaptive layer-wise importance sampling framework. GRASS utilizes mean gradient norms as a task-aware and training-stage-aware metric for estimating layer importance. Furthermore, GRASS adaptively adjusts layer sampling probabilities through an adaptive training strategy. We also introduce a layer-wise optimizer state offloading mechanism that overlaps computation and communication to further reduce memory usage while maintaining comparable training throughput.
Extensive experiments across multiple models and benchmarks demonstrate that GRASS consistently outperforms state-of-the-art methods, achieving an average accuracy improvement of up to 4.38 points and reducing memory usage by up to 19.97\%.
\end{abstract}

\section{Introduction}
\label{sec:introduction}
Large language models (LLMs) serve as the foundation of modern natural language processing, delivering strong performance across diverse tasks~\citep{brown_language_2020, li_pretrained_2021, zhu_multilingual_2024}. To adapt LLMs for downstream tasks, full-parameter fine-tuning (FFT)~\citep{howard_universal_2018} is widely adopted.
Nevertheless, as model sizes scale, FFT becomes increasingly costly and impractical due to the prohibitive bottleneck of GPU memory~\citep{wang_parameterefficient_2025}.
Parameter-efficient fine-tuning (PEFT) methods overcome this issue by updating only a small subset of parameters. Among these, LoRA~\citep{hu_lora_2021a} has gained widespread popularity because of its effective trade-off between efficiency and model performance.
Despite their efficiency, PEFT methods inevitably yield inferior results to FFT. Recent work shows that LoRA~\citep{hu_lora_2021a}, in particular, suffers from limited representational capacity due to its low-rank parameterization~\citep{ding_delta_2022, lialin_relora_2023}, resulting in degraded performance compared to FFT~\citep{xia_chain_2024}.

Unlike LoRA, layer-wise fine-tuning methods~\citep{zhu_lift_2023, pan_lisa_2024a, yao_layerwise_2024, li_outlierweighed_2025} provide an alternative approach to reducing the memory cost of FFT. These methods avoid low-rank constraints and instead activate and update only a subset of layers during training, thereby reducing memory overhead while preserving the model's full capacity.
However, existing layer-wise strategies rely on static, task-agnostic criteria for layer selection, implicitly assuming that layer importance remains constant across tasks and throughout training.
Empirical evidence indicates that the relative importance of layers varies across downstream tasks and training stages. As a result, static strategies often misalign with training dynamics and yield suboptimal performance on certain benchmarks, as shown in Table~\ref{tab:fft_lisa_math}.

To overcome these limitations, we propose GRASS, a
\underline{gr}adient-based \underline{a}daptive layer-wi\underline{s}e importance \underline{s}ampling framework for memory-efficient LLM fine-tuning. 
First, we pinpoint the limitations of static layer-wise sampling methods on downstream tasks and compare the contribution of different layers to training. Our findings reveal the varying layer-wise importance across tasks and training stages. Driven by these insights, GRASS directly uses dynamic optimization signals rather than static heuristics or task-specific assumptions to guide layer selection.
Specifically, GRASS leverages mean gradient norms (MGN) as a task-aware and training-stage-aware indicator to quantify the contribution of each layer to loss reduction. By periodically measuring layer-wise MGN and dynamically updating sampling probabilities, GRASS adaptively samples and activates a subset of layers for fine-tuning, thereby focusing on the most influential layers at different points during training. This mechanism ensures the full-parameter expressiveness of the model. In addition, we introduce a layer-wise optimizer state offloading mechanism to further improve memory efficiency. By overlapping optimizer state transfers with computation, GRASS maintains comparable training throughput.

The contributions of this work are as follows:
\begin{itemize}
    \item We identify the limitations of the static layer-wise sampling strategy through experiments on different datasets, demonstrating that the importance of each layer varies across tasks and training stages. 
    \item We propose GRASS, a novel gradient-based adaptive layer-wise fine-tuning framework. It assesses the importance of layers based on the mean gradient norms and adopts adaptive layer sampling probabilities to achieve an average performance improvement. 
    \item We further address the memory bottleneck of layer-wise fine-tuning by introducing a layer-wise optimizer state offloading mechanism with overlapping, enabling GRASS to achieve high memory efficiency without sacrificing training throughput.
\end{itemize}

\section{Related Work}
\label{sec:related_work}

\subsection{Parameter-efficient Fine-tuning}
As LLMs scale, FFT becomes increasingly impractical due to memory constraints. PEFT methods address this challenge by updating only a subset of the pre-trained parameters. Representative PEFT approaches can be categorized into: prompt-based methods~\citep{lester_power_2021a, li_prefixtuning_2021, liu_ptuning_2022}, adapter-based methods~\citep{houlsby_parameterefficient_2019b, wang_adamix_2022, he_unified_2022}, and reparameterization-based methods~\citep{hu_lora_2021a, valipour_dylora_2023a, liu_dora_2024}. 
Among them, LoRA~\citep{hu_lora_2021a} is widely adopted due to its simplicity and effectiveness. LoRA introduces two low-rank matrices to approximate the incremental parameters, and merges them into the pre-trained weights during inference, incurring no additional computational overhead. Despite its benefits, the expressiveness of LoRA is limited by the number of its trainable parameters and low-rank parameterization, thus falling short of FFT~\citep{lialin_relora_2023, xia_chain_2024, zhao_galore_2024}.
Then, DoRA~\citep{liu_dora_2024} revisits LoRA from a weight decomposition perspective. By explicitly decoupling weight magnitude and direction to enable learnable magnitude, DoRA enhances the expressiveness of LoRA without compromising its parameter efficiency. However, DoRA still uses a limited number of trainable parameters and requires more memory compared to LoRA.

\begin{table}[t]
\centering
\resizebox{1\linewidth}{!}{
\begin{tabular}{@{}lcccccccl@{}}
\toprule
\textbf{Method} & \textbf{MultiA.} & \textbf{AddSub} & \textbf{GSM8K} & \textbf{AQuA} & \textbf{SingleEq} & \textbf{SVAMP} & \textbf{Avg. $\uparrow$} \\ \midrule
FFT & \textbf{94.43} & \textbf{66.08} & \textbf{47.31} & 18.90 & \textbf{80.51} & \textbf{55.50} & \textbf{60.46} \\
LISA & 91.17 & 62.53 & 42.91 & \textbf{20.08} & 71.65 & 51.10 & 56.57 \\ \bottomrule
\end{tabular}%
}
\caption{Comparison of full-parameter fine-tuning (FFT) and LISA on arithmetic reasoning benchmarks. Although LISA approaches FFT performance on certain tasks, it exhibits notable degradation on others(e.g., GSM8K and SingleEq).}
\label{tab:fft_lisa_math}
\end{table}

\subsection{Static Layer-wise Selective Fine-tuning}
Other works aim to reduce GPU memory cost during fine-tuning by updating only a subset of layers. 
LIFT~\citep{zhu_lift_2023} proposes an extreme approach that updates only one Transformer block per iteration. It shortens the backward depth and reduces optimizer state memory requirements. However, LIFT utilizes a fixed update order (e.g., front-to-end), resulting in the allocation of computational budget to layers with low task relevance. Moreover, LIFT lacks a mechanism for prioritizing more informative layers.
\citet{pan_lisa_2024a} observe a skewed weight-norm distribution in LoRA. Inspired by importance sampling~\citep{zhao_stochastic_2015}, they propose LISA, which uniformly samples a subset of layers to optimize during training. LISA achieves good performance while significantly reducing memory overhead. Nevertheless, the static sampling scheme in LISA inherently struggles to adapt to the varying characteristics across tasks and training stages.
Recent advances further explore improved static or heuristic-driven layer selection strategies. IST~\citep{yao_layerwise_2024} estimates layer importance via response suppression and reinforcement learning. OWS~\citep{li_outlierweighed_2025} adopts outlier-weighted sampling with low-rank gradient projection.
In contrast, GRASS employs gradient-based importance estimation that reflects real-time training dynamics and adaptively updates sampling probabilities. Our approach enables more task-aligned layer sampling and consistently yields improved performance across various models and benchmarks.

\section{Method}
\label{sec:method}
\subsection{Gradient-based Layer-wise Importance Measure}

\begin{figure}[t]
    \centering
    \includegraphics[width=1\linewidth]{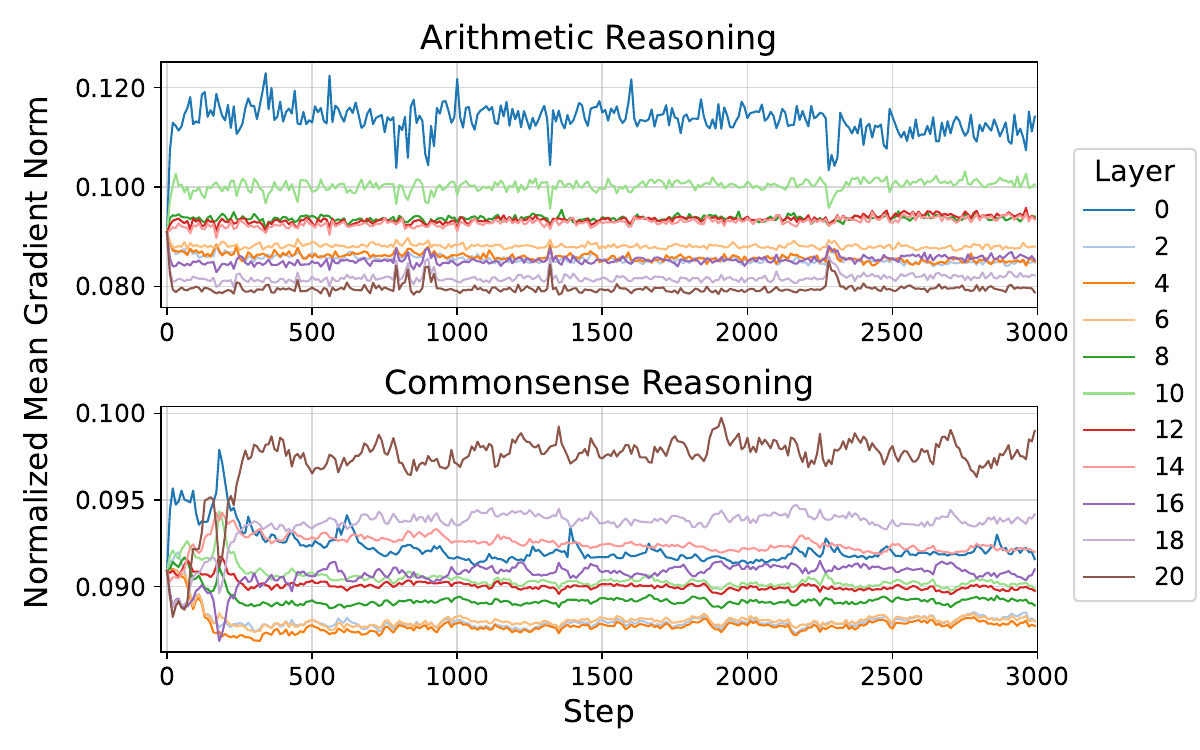}
    \caption{A comparison of normalized layer-wise mean gradient norm across different datasets on TinyLlama.}
    \label{fig:mgn}
\end{figure}

Selecting which layers to update is crucial to layer-wise fine-tuning. Previous methods, such as LISA, estimate layer importance using a static and weight-based strategy, which implicitly assumes that the relative importance of each layer remains constant across tasks and throughout training.
Consequently, it overlooks the fact that layer importance can vary substantially across different downstream tasks and different optimization stages, leading to suboptimal layer selection.
For example, as shown in Table~\ref{tab:fft_lisa_math}, although LISA significantly reduces memory consumption, it underperforms FFT on most arithmetic reasoning benchmarks.
This performance gap suggests that static strategies can misidentify which layers are truly influential for the current task, thereby limiting optimization effectiveness.

To develop an importance measure that more directly reflects the current training dynamics, we turn to the gradient statistics of each layer. Intuitively, gradients encode how sensitively the loss responds to parameter updates. Under the first-order Taylor approximation, we have:
\begin{equation}
  \label{eq:taylor_approximation}
  \Delta \mathcal{L} \approx \langle \nabla_{\theta^{(l)}}\mathcal{L}, \Delta \theta^{(l)} \rangle ~,
\end{equation}
where $\theta^{(l)}$ denotes the parameters of layer $l$, $\Delta \theta^{(l)}$ the parameter update applied to that layer at a training step, and $\Delta \mathcal{L}$ the corresponding change in training loss. The magnitude of the gradient with respect to layer $l$ quantifies the potential impact of updating that layer on the training objective, indicating higher importance for optimization. 

Based on this insight, we define the mean gradient norm (MGN) to measure layer-wise importance by aggregating gradient magnitudes over continuous training steps. Let $\mathbf{g}_t^{(l)}$ denote the gradients of layer $l$ at training step $t$, $N_p^{(l)}$ the number of parameters in that layer, and $T$ the number of continuous training steps. The MGN for layer $l$ within a period composed of $T$ training steps is defined as: 
\begin{equation}
  \label{eq:mgn}
  \mathbf{m}_l(T) = \frac{1}{T}\sum_{t=1}^{T}\sqrt{\frac{1}{N_p^{(l)}}\Arrowvert \mathbf{g}_t^{(l)} \Arrowvert_2^2} ~.
\end{equation}

MGN serves as a task-aware and training-stage-aware indicator of layer importance. In our formulation, layers with higher MGN values tend to be more important, as these layers have a greater impact on loss reduction when updated. Detailed discussion of the validity of MGN as a layer-wise importance measure is provided in Appendix~\ref{sec:validity_of_mgn}.
Figure~\ref{fig:mgn} illustrates the variation of normalized MGN values across layers when fine-tuning TinyLlama on different datasets. We observe significant differences in the relative importance of layers across tasks. Specifically, layer 20 consistently exhibits higher MGN under commonsense reasoning, whereas it is less prominent for arithmetic reasoning. This divergence suggests that layer importance is highly task-dependent and changes continuously as training proceeds.
These observations motivate the use of MGN as the core signal for adaptive layer sampling in GRASS. It allows layer importance to be estimated in a task-aware and training-stage-aware manner, rather than relying on static criteria that ignore the evolving optimization dynamics. Additional analyses conducted on different models further confirm that MGN effectively captures layer-wise differentiation and task-specific patterns (see Appendix~\ref{sec:layer-wise_mgn}).

\subsection{Layer Sampling with Adaptive Update of Sampling Probabilities}

\begin{figure}[t]
    \centering
    \includegraphics[width=1\linewidth]{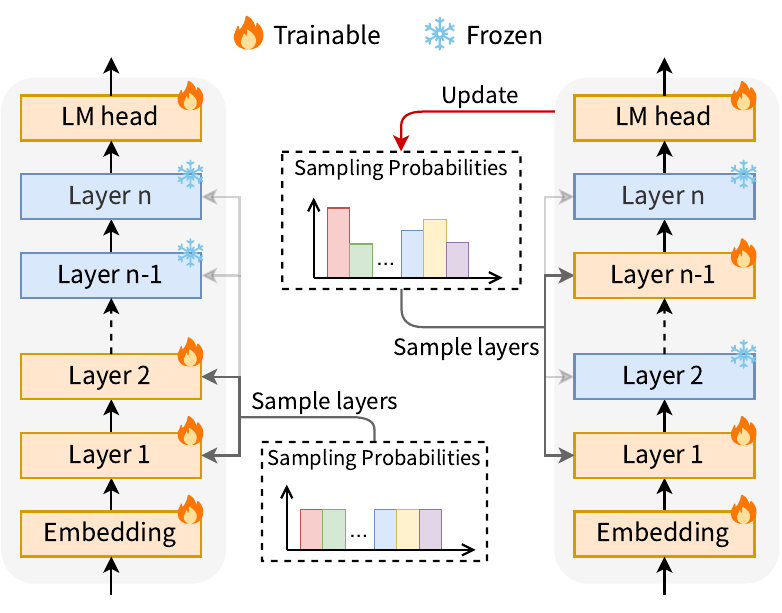}
    \caption{An overview of static layer-wise sampling (Left) and GRASS (Right). GRASS alternates between sampling a subset of model layers for tuning and updating sampling probabilities during training.}
    \label{fig:grass}
\end{figure}

Building upon MGN as a dynamic indicator of layer-wise importance introduces a new challenge: static layer sampling strategies are unable to exploit such time-varying signals. In particular, sampling decisions based on initial statistics can quickly become misaligned with evolving optimization dynamics, resulting in suboptimal performance (see further discussion and analysis in \S~\ref{sec:ablation_study}).
To address this challenge, GRASS dynamically updates both MGN statistics and layer sampling probabilities throughout training.

GRASS proceeds in two stages: a probing phase for initializing layer-wise MGN values, followed by an adaptive fine-tuning phase guided by these MGNs.
At the start of training, GRASS conducts a probing phase for the first $T_p$ steps ($T_p \ll T$) to initialize each layer's MGN.
During probing, standard forward and backward passes are performed as in FFT, but parameter updates are omitted. This stage allows us to collect gradient statistics without incurring the memory overhead of optimizer states. The probing phase yields an initial MGN estimate $\mathbf{m}_l(T_p)$ for each layer, which serves as the starting point for subsequent adaptive sampling.

After probing, GRASS transitions to the adaptive fine-tuning phase, where it periodically samples layers for update. At every probability update interval of $T_u$ training steps, the current MGN values are converted into a probability distribution over layers:
\begin{equation}
  \label{eq:sampling_probabilities}
  \{p^{(l)}\}_{l=1}^{N_L} = \frac{\exp(\mathbf{m}_l(T_u)/\tau)}{\sum_{i=1}^{N_L}\exp(\mathbf{m}_i(T_u)/\tau)} ~,
\end{equation}
where $\tau$ is a temperature hyperparameter controlling the sharpness of the distribution. This softmax-based mapping assigns higher sampling probabilities to layers with larger MGNs, while preserving exploration over less active layers. 
Based on $\{p^{(l)}\}_{l=1}^{N_L}$, GRASS samples $\gamma$ layers out of $N_L$ to update during the subsequent sampling period of length $T_s$, while all other layers remain frozen. The sampled layers in each sampling period are referred to as \emph{trainable} layers, and the remaining layers are considered \emph{frozen}, as shown in Figure~\ref{fig:grass}. During each update interval, only the trainable layers produce gradients while the frozen layers do not. 
We update MGN using an exponential moving average:
\begin{equation}
  \label{eq:update_mgn}
  \mathbf{m}_l(T) = \alpha\mathbf{m}_l(T_u)+(1-\alpha)\mathbf{m}_l(T-T_u) ~,
\end{equation}
where $\alpha$ controls the trade-off between responsiveness to newly observed MGNs and stability of the accumulated importance estimates. Frozen layers retain their previous MGN values during the current update period.
This adaptive mechanism enables GRASS to progressively emphasize layers that consistently exhibit high importance, while also allowing layer importance to evolve with the training stage and downstream task characteristics. 

In summary, GRASS first initializes layer-wise MGN values through a probing stage, then alternates between (i) sampling a small subset of layers for parameter updates and (ii) refreshing their importance estimates based on newly observed gradient statistics.
Unlike existing layer-wise sampling methods that rely on static importance scores (Figure~\ref{fig:grass} (Left)), GRASS dynamically aligns layer selection with the ongoing optimization process by continuously updating sampling probabilities using MGN, as depicted in Figure~\ref{fig:grass} (Right).

\subsection{Layer-wise Optimizer State Offloading}
\label{sec:layer-wise_optimizer_state_offloading}

Layer-wise fine-tuning methods introduce a new memory challenge that differs from LoRA-style approaches, which maintain optimizer states for a fixed and small set of parameters. These methods dynamically activate different subsets of model layers during training. Consequently, the optimizer states associated with all potentially trainable layers need to be preserved throughout training.
A straightforward solution is to keep all the optimizer states in GPU memory. However, this approach incurs prohibitive memory overhead, undermining the memory efficiency benefits of layer-wise fine-tuning. Alternatively, storing optimizer states on CPU reduces GPU memory consumption but introduces significant communication latency due to frequent data transfers between CPU and GPU. This dilemma between memory and throughput poses a challenge in layer-wise fine-tuning methods.  

\begin{figure}[t]
    \centering
    \includegraphics[width=1\linewidth]{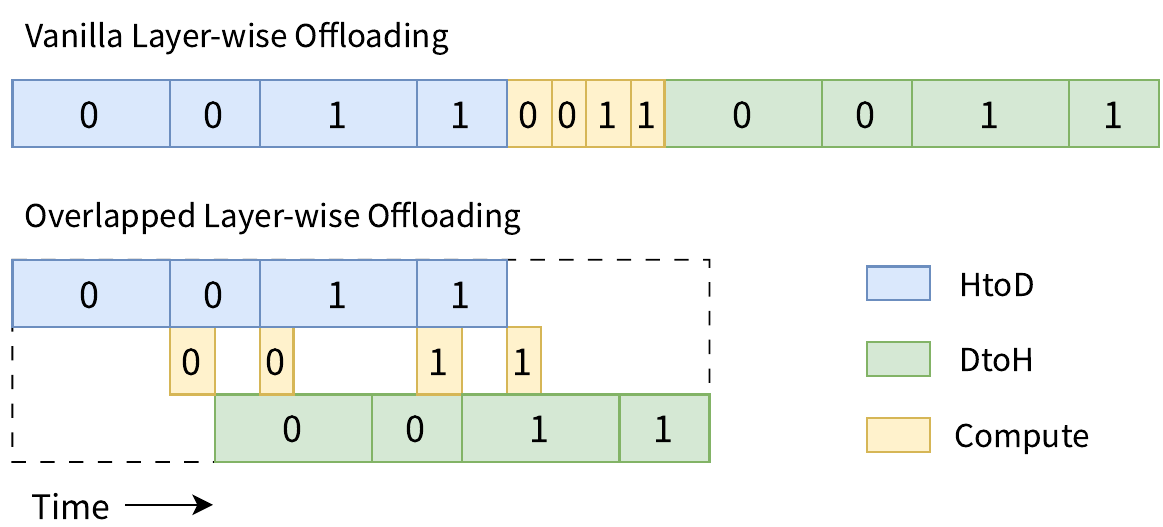}
    \caption{A comparison of vanilla and overlapped layer-wise optimizer state offloading. HtoD refers to host to device, while DtoH represents device to host. Each block with index $i$ stands for a parameter in layer $i$.}
    \label{fig:overlapping}
\end{figure}

GRASS addresses this dilemma by exploiting a key observation: at any training step, only a small subset of layers (i.e., $\gamma \ll N_L$) are active and require optimizer updates. Therefore, we adopt a layer-wise optimizer state offloading strategy, where optimizer states are managed at the granularity of individual layers.
Specifically, at any given time, GPU memory only holds the optimizer states of one active layer that is currently undergoing updates, while the states of remaining layers are stored in CPU memory. To mitigate communication overhead, GRASS asynchronously prefetches optimizer states to perform updates, then puts them back to the CPU immediately after the updates layer by layer. By overlapping data transfers with computation, this approach reduces idle time during optimizer updates, thereby increasing training throughput (as shown in Figure~\ref{fig:overlapping}), while minimizing the optimizer state memory footprint on the GPU. 
We provide quantitative analysis of the benefits of this technique in \S~\ref{sec:memory_efficiency}.

\section{Experiments}
\label{sec:experiments}

\subsection{Arithmetic Reasoning}
\label{arithmetic_reasoning}

\begin{table*}[t]
\centering
\resizebox{0.80\linewidth}{!}{%
\begin{NiceTabular}{@{}clccccccl@{}}
\toprule
\textbf{Model} & \multicolumn{1}{c}{\textbf{Method}} & \multicolumn{1}{c}{\textbf{MultiArith}} & \multicolumn{1}{c}{\textbf{AddSub}} & \multicolumn{1}{c}{\textbf{GSM8K}} & \multicolumn{1}{c}{\textbf{AQuA}} & \multicolumn{1}{c}{\textbf{SingleEq}} & \multicolumn{1}{c}{\textbf{SVAMP}} & \multicolumn{1}{c}{\textbf{Avg. $\uparrow$}} \\ \midrule
\multirow{6}{*}{TinyLlama}
 & \rowcolor{gray} FFT & 64.17 & 40.76 & 15.16 & 13.78 & 42.92 & 24.10 & 33.48 \\
 & LoRA\textsubscript{r=128} & 61.17 & 25.32 & 15.16 & 12.20 & 38.19 & 27.00 & 29.84 \\
 & LoRA\textsubscript{r=256} & 65.50 & 30.89 & 17.44 & 12.60 & 37.59 & \underline{28.20} & 32.04 \\
 & DoRA & \underline{67.83} & 34.68 & \textbf{18.35} & \underline{12.99} & 36.81 & \textbf{29.50} & 33.36 \\
 & LISA & 65.00 & \textbf{38.04} & \underline{17.74} & \underline{12.99} & \textbf{43.11} & 24.90 & \underline{33.63} \\
 & \rowcolor{grassgreen} \textbf{GRASS} & \textbf{68.00} & \underline{35.19} & 17.13 & \textbf{15.16} & \underline{42.52} & 27.30 & \textbf{34.22} \\ \midrule
\multirow{6}{*}{Gemma-2B}
 & \rowcolor{gray} FFT & 86.67 & 66.84 & 42.53 & 32.68 & 80.12 & 52.10 & 60.16 \\
 & LoRA\textsubscript{r=128} & 91.50 & \underline{66.84} & 42.61 & \underline{25.20} & 77.36 & 49.00 & 58.75 \\
 & LoRA\textsubscript{r=256} & \underline{92.50} & 65.06 & 42.53 & 23.23 & \textbf{79.33} & \underline{52.30} & 59.16 \\
 & DoRA & 92.17 & \textbf{67.09} & \underline{42.68} & 23.62 & \underline{78.94} & 50.90 & \underline{59.23} \\
 & LISA & 90.17 & 62.53 & 40.18 & 21.26 & 75.00 & 49.60 & 56.46 \\
 & \rowcolor{grassgreen} \textbf{GRASS} & \textbf{93.50} & 66.33 & \textbf{43.06} & \textbf{29.13} & 78.35 & \textbf{53.50} & \textbf{60.65} \\ \midrule
\multirow{6}{*}{LLaMA2-7B}
 & \rowcolor{gray} FFT & 94.43 & 66.08 & 47.31 & 18.90 & 80.51 & 55.50 & 60.46 \\
 & LoRA\textsubscript{r=128} & 90.50 & 62.28 & 46.85 & 19.68 & \textbf{78.74} & 54.90 & 58.83 \\
 & LoRA\textsubscript{r=256} & 90.50 & \underline{63.04} & \underline{46.87} & 19.69 & \underline{78.54} & \textbf{56.00} & 59.11 \\
 & DoRA & \textbf{93.16} & 63.03 & \textbf{48.52} & \underline{21.65} & 75.59 & 53.40 & \underline{59.23} \\
 & LISA & 91.17 & 62.53 & 42.91 & 20.08 & 71.65 & 51.10 & 56.57 \\
 & \rowcolor{grassgreen} \textbf{GRASS} & \underline{92.00} & \textbf{67.09} & 42.15 & \textbf{23.62} & 77.56 & \underline{55.10} & \textbf{59.59} \\ \bottomrule
\end{NiceTabular}%
}
\caption{Accuracy comparison of multiple LLMs and PEFT methods on six math reasoning benchmarks. The best and second best results are marked in \textbf{bold} and \underline{underlined}, respectively.}
\label{tab:math}
\end{table*}

\paragraph{Settings}
To validate the effectiveness of GRASS, we compare it with various PEFT methods across three language models of different scales on arithmetic reasoning tasks. Specifically, we conduct experiments on TinyLlama~\citep{zhang_tinyllama_2024}, Gemma-2B~\citep{team_gemma_2024}, and LLaMA2-7B~\citep{touvron_llama_2023}, covering parameters counts from 1B to 7B. Each model is first fine-tuned with the math dataset collected by~\citet{hu_llmadapters_2023}, and subsequently evaluated on six arithmetic reasoning benchmarks, including MultiArith~\citep{roy_solving_2015}, AddSub~\citep{hosseini_learning_2014}, GSM8K~\citep{cobbe_training_2021}, AQuA~\citep{ling_program_2017}, SingleEq~\citep{koncel-kedziorski_parsing_2015}, and SVAMP~\citep{patel_are_2021}. Additional experimental details are provided in Appendix~\ref{sec:experimental_details}.

\paragraph{Results}
Table~\ref{tab:math} reports the performance of different PEFT methods on arithmetic reasoning benchmarks across three model scales. Overall, GRASS yields competitive results across all models, surpassing state-of-the-art methods on most benchmarks, indicating its robustness and effectiveness. 
Specifically, GRASS consistently achieves higher average accuracy than both LoRA and DoRA. This result suggests that adaptive layer-wise updates offer greater representational flexibility compared to low-rank update methods.
Compared to LISA, GRASS provides more consistent improvements across tasks and model scales. Although LISA performs well on certain benchmarks (e.g., AddSub and SingleEq on TinyLlama), its static sampling strategy can lead to performance declines on other tasks. In contrast, GRASS leverages MGN to dynamically adjust layer sampling probabilities during training, capturing both task-dependent and stage-dependent layer importance. This adaptive assessment allows GRASS to accommodate diverse tasks, leading to more balanced performance gains. Notably, GRASS outperforms FFT on both TinyLlama and Gemma-2B, and achieves an average accuracy improvement of up to 4.38 points over LoRA\textsubscript{r=128}. These results suggest that adaptive layer-wise fine-tuning may serve as an implicit regularizer, consistent with findings from prior works~\citep{pan_lisa_2024a}.

\subsection{Commonsense Reasoning}
\label{commonsense_reasoning}

\begin{table*}[t]
\centering
\resizebox{0.87\linewidth}{!}{%
\begin{NiceTabular}{@{}clccccccccl@{}}
\toprule
\textbf{Model} & \multicolumn{1}{c}{\textbf{Method}} & \textbf{BoolQ} & \textbf{PIQA} & \textbf{SIQA} & \textbf{HellaS.} & \textbf{WinoG.} & \textbf{ARC-e} & \textbf{ARC-c} & \textbf{OBQA} & \multicolumn{1}{c}{\textbf{Avg. $\uparrow$}} \\ \midrule
\multirow{6}{*}{TinyLlama} 
 & \rowcolor{gray} FFT & 61.90 & 51.25 & 36.64 & 29.72 & 50.43 & 27.95 & 25.85 & 31.00 & 39.34 \\
 & LoRA\textsubscript{r=128} & 58.35 & \textbf{52.50} & 34.34 & \underline{26.17} & 50.75 & 25.04 & 25.68 & 24.80 & 37.20 \\
 & LoRA\textsubscript{r=256} & 58.69 & 50.76 & 33.52 & 25.59 & \textbf{51.70} & \underline{26.64} & \textbf{27.82} & 26.20 & \underline{37.62} \\
 & DoRA & 57.71 & \underline{51.74} & \underline{34.39} & 25.98 & \underline{50.83} & 26.01 & \underline{27.05} & 26.40 & 37.51 \\
 & LISA & \underline{60.18} & 49.05 & 33.78 & 25.77 & 49.49 & 26.22 & 25.43 & \underline{28.80} & 37.34 \\
 & \rowcolor{grassgreen} \textbf{GRASS} & \textbf{61.29} & \underline{51.74} & \textbf{34.75} & \textbf{27.01} & 50.12 & \textbf{27.74} & 26.45 & \textbf{29.60} & \textbf{38.59} \\ \midrule
\multirow{6}{*}{Gemma-2B}
 & \rowcolor{gray} FFT & 64.46 & 76.82 & 66.02 & 73.61 & 60.46 & 80.01 & 64.68 & 70.00 & 69.51 \\
 & LoRA\textsubscript{r=128} & 60.21 & 73.29 & 61.59 & 68.65 & 59.43 & 76.89 & 60.67 & 67.60 & 66.04 \\
 & LoRA\textsubscript{r=256} & 60.40 & 73.45 & 62.54 & 68.07 & \underline{61.17} & 78.28 & 60.32 & 66.40 & 66.33 \\
 & DoRA & 61.68 & 73.50 & 62.74 & 68.56 & 59.75 & 77.65 & 60.67 & 66.00 & 66.32 \\
 & LISA & \underline{62.63} & \textbf{78.18} & \underline{65.61} & \underline{72.87} & 60.77 & \underline{78.91} & \textbf{62.03} & \underline{68.00} & \underline{68.63} \\
 & \rowcolor{grassgreen} \textbf{GRASS} & \textbf{63.79} & \underline{76.01} & \textbf{66.73} & \textbf{76.10} & \textbf{62.27} & \textbf{79.29} & \underline{61.60} & \textbf{68.60} & \textbf{69.30} \\ \midrule
\multirow{8}{*}{LLaMA2-7B}
 & \rowcolor{gray} FFT & 71.16 & 80.96 & 75.13 & 85.88 & 71.35 & 86.28 & 72.27 & 79.40 & 77.80 \\
 & LoRA\textsubscript{r=128} & 66.70 & 78.35 & 73.85 & 82.90 & 70.24 & 84.85 & 68.03 & 77.40 & 75.29 \\
 & LoRA\textsubscript{r=256} & 67.40 & 78.35 & 74.05 & 83.10 & 71.59 & \underline{84.97} & \textbf{69.80} & \underline{78.00} & 75.91 \\
 & DoRA & 68.53 & 79.65 & 73.13 & 83.37 & \textbf{72.45} & \textbf{85.27} & \underline{69.20} & 77.60 & \underline{76.15} \\
 & LISA & \underline{68.93} & \underline{80.06} & 72.42 & 83.51 & 68.59 & 83.59 & 67.66 & 75.60 & 75.05 \\
 & IST & 68.20 & 78.73 & 73.08 & \textbf{84.19} & \underline{71.64} & 82.87 & 68.98 & 76.60 & 75.54 \\
 & OWS & \textbf{69.73} & 79.92 & \textbf{74.96} & 82.93 & 70.43 & 82.70 & 66.21 & 77.80 & 75.59 \\
 & \rowcolor{grassgreen} \textbf{GRASS} & 68.87 & \textbf{80.09} & \underline{74.31} & \underline{83.71} & 71.35 & 84.72 & 69.11 & \textbf{78.20} & \textbf{76.30} \\ \bottomrule
\end{NiceTabular}%
}
\caption{Accuracy comparison of multiple LLMs and PEFT methods on eight commonsense reasoning  benchmarks. The best and second best results are marked in \textbf{bold} and \underline{underlined}, respectively.}
\label{tab:commonsense}
\end{table*}

\paragraph{Settings}
To further demonstrate the generalization of GRASS across diverse tasks, we conduct fine-tuning experiments on commonsense reasoning. 
Following the methodology of~\citet{hu_llmadapters_2023}, we employ a training dataset comprising eight tasks and conduct evaluations on the individual test dataset for each task, respectively. The commonsense reasoning tasks include BoolQ~\citep{clark_boolq_2019}, PIQA~\citep{bisk_piqa_2020}, SIQA~\citep{sap_social_2019}, HellaSwag~\citep{zellers_hellaswag_2019}, WinoGrande~\citep{sakaguchi_winogrande_2021}, ARC-c/e~\citep{clark_think_2018}, and OBQA~\citep{mihaylov_can_2018}. 

\paragraph{Results}
As shown in Table~\ref{tab:commonsense}, GRASS delivers consistent improvements across all three model scales on commonsense reasoning tasks.
On TinyLlama, GRASS improves the average accuracy to 38.59\%, outperforming LoRA, DoRA, and LISA. This improvement demonstrates that adaptive layer selection remains advantageous even for smaller models with limited parameter capacity, where static or task-agnostic update strategies may waste optimization effort on less relevant layers.
For LLaMA-2-7B, GRASS achieves the closest performance to FFT (76.30\% vs. 77.80\%), outperforming all other PEFT methods. 
These results highlight that GRASS can adapt its sampling policy based on gradients and focus on layers that exhibit higher task-specific importance, yielding superior performance. 
Overall, GRASS demonstrates strong generalization beyond arithmetic reasoning and provides robust improvements on various commonsense reasoning benchmarks.

\subsection{Memory Efficiency}
\label{sec:memory_efficiency}

\begin{table}[t]
\centering
\resizebox{\linewidth}{!}{%
\begin{tabular}{@{}c|r|rr|r|rr|rr@{}}
\toprule
\textbf{} & \multicolumn{1}{c|}{\textbf{FFT}} & \multicolumn{2}{c|}{\textbf{LoRA}} & \multicolumn{1}{c|}{\textbf{DoRA}} & \multicolumn{2}{c|}{\textbf{LISA}} & \multicolumn{2}{c}{\textbf{GRASS}} \\ \midrule
\textbf{Model} & \multicolumn{1}{c|}{-} & \multicolumn{1}{c}{r=128} & \multicolumn{1}{c|}{r=256} & \multicolumn{1}{c|}{-} & \multicolumn{1}{c}{$\gamma$=2} & \multicolumn{1}{c|}{$\gamma$=4} & \multicolumn{1}{c}{$\gamma$=2} & \multicolumn{1}{c}{$\gamma$=4} \\ \midrule
TinyLlama & 8.76 & 4.71 & 5.03 & 5.71 & 4.93 & 5.31 & 4.49 & 4.55 \\
Gemma-2B & 21.19 & 11.26 & 11.81 & 11.64 & 13.33 & 14.27 & 12.45 & 12.46 \\
LLaMA2-7B & 51.32 & 19.97 & 20.86 & 23.23 & 20.68 & 22.31 & 19.08 & 19.22 \\ \bottomrule
\end{tabular}%
}
\caption{Comparison of peak GPU memory consumption (GB) for different models and fine-tuning methods. Batch size and sequence length are set to 1 and 1024, respectively.}
\label{tab:peak_mem}
\end{table}

We conduct experiments on the peak GPU memory consumption of different fine-tuning methods across several scenarios to showcase the memory efficiency of GRASS.

\begin{figure}[t]
  \includegraphics[width=0.529\linewidth]{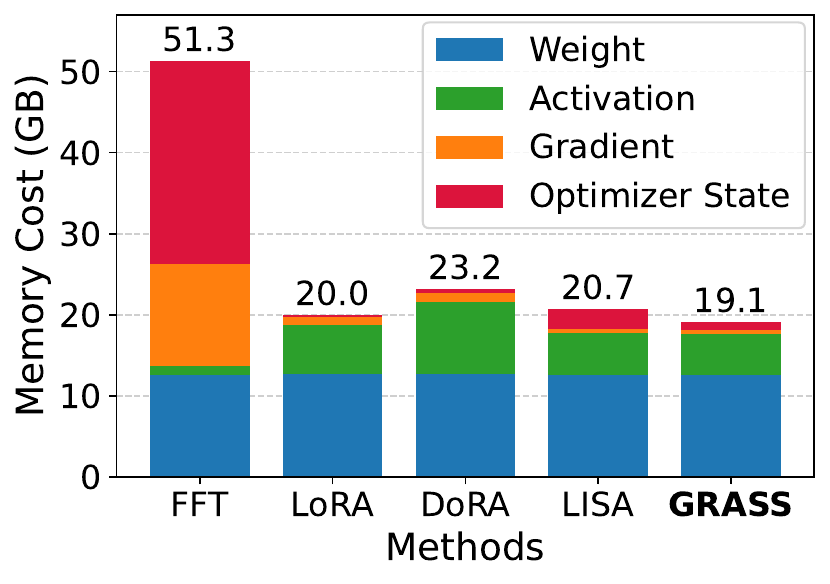} \hfill
  \includegraphics[width=0.461\linewidth]{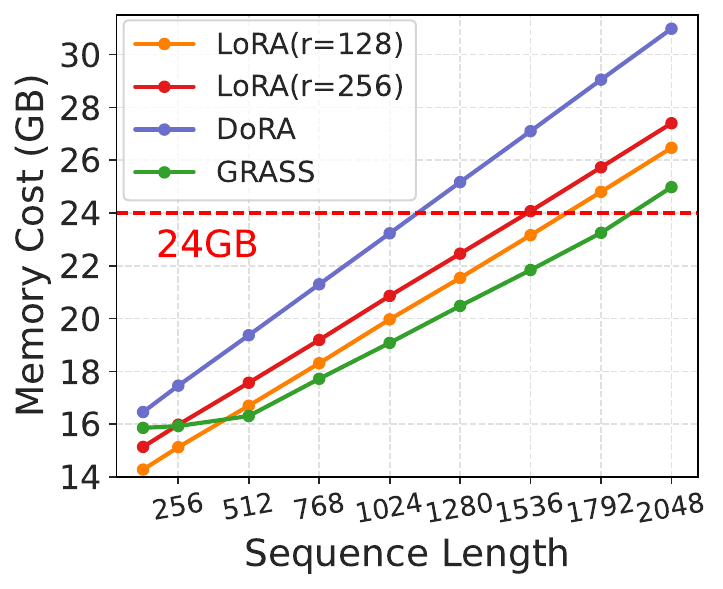}
  \caption {(Left) Peak memory consumption of fine-tuning LLaMA2-7B with different methods. (Right) Memory consumption of fine-tuning LLaMA2-7B across different sequence lengths.}
  \label{fig:memory}
\end{figure}

Table~\ref{tab:peak_mem} presents the peak GPU memory usage across different model sizes. Compared to reparameterization-based methods such as LoRA and DoRA, GRASS consistently achieves a lower or comparable memory footprint. This advantage stems from a reduction in activation memory, as depicted in Figure~\ref{fig:memory} (Left). Regarding LISA, GRASS effectively reduces optimizer state memory consumption. Moreover, the memory usage of GRASS exhibits only a marginal increase as the number of active layers $\gamma$ grows, demonstrating a significant reduction in memory growth, from 1.63 GB to just 0.14 GB. This reduction can be attributed to the layer-wise optimizer states offloading technique discussed in \S~\ref{sec:layer-wise_optimizer_state_offloading}. 

To investigate how memory usage scales with input sequence length, we conducted experiments on LLaMA2-7B with a fixed batch size of 1. As shown in Figure~\ref{fig:memory} (Right), DoRA consistently consumes more memory than GRASS as the sequence length grows, and GRASS maintains the lowest memory usage under most scenarios (sequence length $\geq$ 512). 
Notably, when the sequence length reaches 1792, the memory consumption of LoRA (r=128/256) and DoRA exceeds the 24~GB limit, whereas GRASS remains below this threshold at 23.25~GB. 
These results highlight that GRASS achieves substantial reductions in peak memory usage while significantly enhancing long-context scalability, thus enabling memory-efficient training under hardware constraints where FFT and even some PEFT methods become infeasible.

\subsection{Ablation Study}
\label{sec:ablation_study}

\paragraph{Static strategy}

\begin{table*}[t]
\centering
\resizebox{0.78\linewidth}{!}{%
\begin{NiceTabular}{@{}clccccccl@{}}
\toprule
\textbf{Model} & \textbf{Method} & \textbf{MultiArith} & \textbf{AddSub} & \textbf{GSM8K} & \textbf{AQuA} & \textbf{SingleEq} & \textbf{SVAMP} & \textbf{Avg. $\uparrow$} \\ \midrule
\multirow{2}{*}{TinyLlama} & GRASS* & 54.50 & \textbf{39.24} & 13.12 & \textbf{15.75} & 40.75 & 20.80 & 30.69 \\
 & \rowcolor{grassgreen} GRASS & \textbf{68.00} & 35.19 & \textbf{17.13} & 15.16 & \textbf{42.52} & \textbf{27.30} & \textbf{34.22} \\ \midrule
\multirow{2}{*}{Gemma-2B} & GRASS* & 91.00 & 63.80 & 41.40 & \textbf{31.10} & 76.77 & 49.90 & 59.00 \\
 & \rowcolor{grassgreen} GRASS & \textbf{93.50} & \textbf{66.33} & \textbf{43.06} & 29.13 & \textbf{78.35} & \textbf{53.50} & \textbf{60.65} \\ \midrule
\multirow{2}{*}{LLaMA2-7B} & GRASS* & 91.83 & 61.52 & 41.93 & 13.78 & 77.56 & 52.90 & 56.59 \\
 & \rowcolor{grassgreen} GRASS & \textbf{92.00} & \textbf{67.09} & \textbf{42.15} & \textbf{23.62} & 77.56 & \textbf{55.10} & \textbf{59.59} \\ \bottomrule
\end{NiceTabular}%
}
\caption{Comparison of GRASS and its variant without adaptive update of sampling probabilities (GRASS*) on arithmetic reasoning tasks.}
\label{tab:ablation_static}
\end{table*}

We demonstrate the accuracy gains of adaptive layer sampling by comparing GRASS to a static variant (GRASS*), where layer sampling probabilities are initialized using the MGN obtained in the probing phase and remain fixed throughout training. As shown in Table~\ref{tab:ablation_static}, the results of arithmetic reasoning tasks show that adaptive sampling probability updates have brought comparable or significant accuracy improvements across different tasks and models. These results indicate that layer importance is not only task-dependent but also stage-dependent. 
While the initial probing phase provides a reasonable starting point, fixing the sampling distribution fails to adapt to changing gradient patterns during the optimization process. 
By continuously updating MGN and sampling probabilities, GRASS dynamically aligns layer updates with the current training stage, leading to improved generalization and accuracy.

\paragraph{Hyperparameters}

\begin{table*}[t]
\centering
\resizebox{0.96\linewidth}{!}{%
\begin{tabular}{@{}c|ccccc|ccccc|ccccc@{}}
\toprule
$\gamma$ & \multicolumn{5}{c|}{2} & \multicolumn{5}{c|}{4} & \multicolumn{5}{c}{8} \\ \midrule
$T_s$ & 5 & 10 & 25 & 50 & 100 & 5 & 10 & 25 & 50 & 100 & 5 & 10 & 25 & 50 & 100 \\ \midrule
Avg. & 57.68 & 57.12 & \textbf{59.59} & 58.40 & 58.85 & 58.13 & 59.64 & 60.34 & \textbf{60.37} & 59.32 & 62.22 & 61.87 & 62.32 & \textbf{62.46} & 60.30 \\ \bottomrule
\end{tabular}%
}
\caption{Comparison average accuracy of applying different GRASS configurations on arithmetic reasoning tasks.}
\label{tab:ablation_hyperparams}
\end{table*}

We investigate various combinations of the two key hyperparameters of GRASS: the sampling period $T_s$, which controls how frequently layers are resampled, and the number of active layers $\gamma$, which determines how many layers are updated in each sampling period. Experiments are conducted with LLaMA2-7B, and all settings adopt a learning rate of $\eta=3\times10^{-5}$. Table~\ref{tab:ablation_hyperparams} reports the average accuracy on arithmetic reasoning tasks under different configurations. 
We find that increasing the number of active layers consistently improves performance.
Such results suggest that allocating a larger update budget allows GRASS to approximate or even surpass FFT, while remaining memory-efficient. 
As for the sampling period, $T_s$ regulates the frequency of layer switching and the update of sampling probabilities. A small $T_s$ (e.g., $T_s=5$) leads to inferior performance, likely due to unstable sampling caused by rapidly changing gradient estimates. Conversely, overly large $T_s$ delays the timely update of layer-wise importance. Across all settings, intermediate sampling periods ($T_s\in\{25,50\}$) consistently yield the best results. Overall, these results suggest that \emph{a moderate sampling period combined with a sufficient number of active layers enhances the performance of GRASS.}
In addition to $T_s$ and $\gamma$, GRASS involves a probing phase used for initializing MGN statistics. We find that GRASS is relatively insensitive to the choice of $T_p$ within a reasonable range, and provide a detailed analysis in Appendix~\ref{sec:sensitivity_to_probing_phase_length}.

\paragraph{Random seed}

\begin{table}[t]
\centering
\resizebox{0.7\linewidth}{!}{%
\begin{tabular}{@{}c|ccc@{}}
\toprule
\textbf{Model} & Seed 1 & Seed 2 & Seed 3 \\ \midrule
TinyLlama & 34.36 & 34.22 & 34.12 \\ \midrule
Gemma-2B & 60.64 & 60.65 & 60.85 \\ \midrule
LLaMA2-7B & 59.45 & 59.59 & 59.37 \\ \bottomrule
\end{tabular}%
}
\caption{Average accuracy on arithmetic reasoning tasks with varying random seeds.}
\label{tab:seed}
\end{table}

Since GRASS involves stochastic layer sampling, we further examine its robustness with respect to different random seeds. Table~\ref{tab:seed} reports the average accuracy on arithmetic reasoning tasks obtained by fine-tuning models with different random seeds. As shown, GRASS exhibits consistent performance across all model scales. The variation across seeds is minimal, with the maximum absolute difference being 0.24 for TinyLlama, 0.21 for Gemma-2B, and 0.22 for LLaMA2-7B. 
These results indicate that the stochasticity introduced by layer sampling does not lead to unstable training outcomes. Overall, these findings confirm the stability of GRASS across random seeds, further supporting its practical reliability.

\paragraph{Throughput}

\begin{figure}[t]
    \centering
    \includegraphics[width=0.90\linewidth]{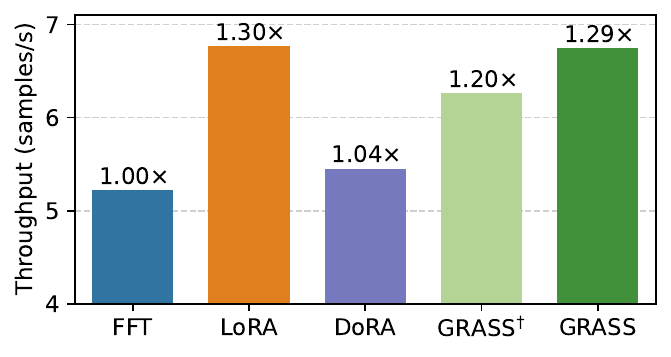}
    \caption{Training throughput of different methods on LLaMA2-7B. $\text{GRASS}^{\dag}$ represents a variant of GRASS that does not use overlap.}
    \label{fig:throughput}
\end{figure}

To demonstrate the training throughput trade-off caused by initializing optimizer states in CPU memory, as well as the enhancement provided by the overlap technique, we conduct experiments on LLaMA2-7B. The batch size and the input sequence length are set to 4 and 1024, respectively. These results are shown in Figure~\ref{fig:throughput}. The results indicate that the overlap technique yields a throughput increase of 1.08$\times$ for GRASS, achieving a throughput comparable to LoRA, and all methods exhibit higher training throughput than FFT.

\paragraph{Computational overhead}

To quantify the additional computational costs introduced by GRASS, we provide a detailed runtime breakdown across three model scales.
As shown in Table \ref{tab:computational_overhead}, the total overhead of GRASS decreases significantly as model size increases. For example, on LLaMA2-7B, the combined overhead of the probing phase and periodic MGN computation constitutes only 2.01\% of the total training time. This reduction occurs because the dominant forward and backward pass computations scale linearly with model size, whereas the computational overhead of GRASS remains nearly constant. These results indicate that the additional cost introduced by probing and periodic MGN computation remains small, approximately 2-6\% of total training time. Considering the memory savings, this overhead does not diminish the practical advantages of GRASS.

\begin{table}[]
\centering
\resizebox{\columnwidth}{!}{%
\begin{tabular}{@{}cccc@{}}
\toprule
\textbf{Model} & \textbf{Probing phase} & \textbf{MGN update} & \textbf{Total training time} \\ \midrule
Tinyllama & 34 (3.24\%) & 32 (3.09\%) & 1050 (100\%) \\ \midrule
Gemma-2B & 30 (0.99\%) & 42 (1.37\%) & 3036 (100\%) \\ \midrule
Llama2-7B & 54 (0.95\%) & 60 (1.06\%) & 5708 (100\%) \\ \bottomrule
\end{tabular}%
}
\caption{Time breakdown of the additional runtime introduced by GRASS on the arithmetic reasoning dataset. Batch size and sequence length are set to 4 and 1024, respectively.}
\label{tab:computational_overhead}
\end{table}

\section{Conclusion}
\label{sec:conclusion}

In this paper, we propose GRASS, a gradient-based adaptive layer-wise importance sampling framework. It dynamically samples and updates influential layers during training based on mean gradient norms. 
By leveraging dynamically estimated layer importance rather than static heuristics, GRASS improves the performance of layer-wise fine-tuning methods.
To reduce the memory of accumulated optimizer states, we introduce layer-wise optimizer state offloading, enabling GRASS to obtain comparable throughput with state-of-the-art methods. Extensive experiments over multiple models and tasks demonstrate that GRASS consistently outperforms PEFT baselines while maintaining matching or lower memory consumption. 

\section*{Limitations}

Despite its effectiveness, GRASS has several limitations that need further investigation. Firstly, it relies on gradient-based statistics to estimate layer importance, which introduces additional computation during training. Nevertheless, the relative overhead decreases as model scale increases, indicating favorable amortization behavior in larger-scale settings.
Additionally, our evaluation focuses on decoder-only language models and widely used NLP benchmarks. The applicability of GRASS to other architectures and modalities remains an interesting direction for future work.




\bibliography{custom}

\clearpage

\appendix

\section{Experimental Details}
\label{sec:experimental_details}
In this section, we give more details about the experiment settings. All experiments are conducted using 2 NVIDIA H100 80G GPUs. For all accuracy-related experiments, we report the average results over three independent runs.

\subsection{Hyperparameters}
We provide detailed hyperparameter settings of arithmetic reasoning and commonsense reasoning tasks in Table~\ref{tab:hyperparameter_math&cs}. Notably, probing phase last for 150 steps ($T_p=150$) in all experiments.

\begin{table*}[t]
\centering
\resizebox{0.9\linewidth}{!}{%
\begin{tabular}{@{}cc|ccc|ccc@{}}
\toprule
\multicolumn{2}{c|}{\multirow{2}{*}{\textbf{Hyperparameter}}} & \multicolumn{3}{c|}{\textbf{Arithmetic Reasoning}} & \multicolumn{3}{c}{\textbf{Commonsense Reasoning}} \\
\multicolumn{2}{c|}{} & TinyLlama & Gemma-2B & LLaMA2-7B & TinyLlama & Gemma-2B & LLaMA2-7B \\ \midrule
\multirow{2.34}{*}{FFT} & Learning rate & \multicolumn{3}{c|}{1e-5} & \multicolumn{3}{c}{1e-5} \\ \cmidrule(l){2-8} 
 & Batch size & 8 & 4 & 4 & 8 & 4 & 4 \\ \midrule
\multirow{5}{*}{LoRA} & Learning rate & \multicolumn{3}{c|}{1e-4} & 5e-5 & 2e-5 & 2e-5 \\ \cmidrule(l){2-8} 
 & Batch size & 8 & 4 & 4 & 8 & 4 & 4 \\ \cmidrule(l){2-8} 
 & Rank & \multicolumn{3}{c|}{128/256} & \multicolumn{3}{c}{128/256} \\ \cmidrule(l){2-8} 
 & $\alpha$ & \multicolumn{3}{c|}{256/512} & \multicolumn{3}{c}{256/512} \\ \midrule
\multirow{5}{*}{DoRA} & Learning rate & \multicolumn{3}{c|}{1e-4} & 5e-5 & 2e-5 & 2e-5 \\ \cmidrule(l){2-8} 
 & Batch size & 8 & 4 & 4 & 8 & 4 & 4 \\ \cmidrule(l){2-8} 
 & Rank & \multicolumn{3}{c|}{128} & \multicolumn{3}{c}{128} \\ \cmidrule(l){2-8} 
 & $\alpha$ & \multicolumn{3}{c|}{256} & \multicolumn{3}{c}{256} \\ \midrule
\multirow{5}{*}{LISA} & Learning rate & \multicolumn{3}{c|}{5e-5} & 5e-5 & 3e-5 & 2e-5 \\ \cmidrule(l){2-8} 
 & Batch size & 8 & 4 & 4 & 8 & 4 & 4 \\ \cmidrule(l){2-8} 
 & $\gamma$ & \multicolumn{3}{c|}{2} & \multicolumn{3}{c}{2} \\ \cmidrule(l){2-8} 
 & $T_s$ & \multicolumn{3}{c|}{10} & \multicolumn{3}{c}{10} \\ \midrule
\multirow{5}{*}{GRASS} & Learning rate & \multicolumn{3}{c|}{3e-5} & 3e-5 & 3e-5 & 2e-5 \\ \cmidrule(l){2-8} 
 & Batch size & 8 & 4 & 4 & 8 & 4 & 4 \\ \cmidrule(l){2-8} 
 & $\gamma$ & \multicolumn{3}{c|}{2} & \multicolumn{3}{c}{2} \\ \cmidrule(l){2-8} 
 & $T_s$ & \multicolumn{3}{c|}{25} & \multicolumn{3}{c}{25} \\ \bottomrule
\end{tabular}%
}
\caption{Hyperparameter configurations of different fine-tuning methods for arithmetic reasoning and common reasoning tasks.}
\label{tab:hyperparameter_math&cs}
\end{table*}

\section{Validity of MGN as a Layer-wise Importance Measure}
\label{sec:validity_of_mgn}

In this section we discuss the rationale of using mean gradient norm (MGN) as a layer-wise importance indicator in GRASS. 

\subsection{Optimization-aligned Proxy}
Gradients provide a direct signal of how sensitively the loss responds to parameter updates. For layer $l$ with parameters $\theta^{(l)}$, a small update $\Delta\theta^{(l)}$ changes the loss approximately via a first-order Taylor expansion:
\begin{equation}
\Delta \mathcal{L} \approx \langle \nabla_{\theta^{(l)}} \mathcal{L}, \Delta \theta^{(l)} \rangle ~, \tag{\ref{eq:taylor_approximation}}
\end{equation}
which implies that the magnitude of $\nabla_{\theta^{(l)}} \mathcal{L}$ reflects the potential impact of updating that layer on the training objective.

Motivated by this observation, we use the gradient norm as a local indicator of a layer’s sensitivity to the loss, reflecting how small parameter perturbations at the current training stage affect the objective.
To ensure comparability across layers of different sizes, we normalize gradient magnitudes by the number of parameters in each layer.
Aggregating these normalized gradient norms over multiple training steps yields the mean gradient norm (MGN), i.e.
\begin{equation}
  \mathbf{m}_l(T) = \frac{1}{T}\sum_{t=1}^{T}\sqrt{\frac{1}{N_p^{(l)}}\Arrowvert \mathbf{g}_t^{(l)} \Arrowvert_2^2} ~, \tag{\ref{eq:mgn}}
\end{equation}
which reduces stochastic noise and captures persistent optimization signals.

Importantly, MGN provides a relative ranking of layers in terms of their short-term optimization impact, rather than an absolute measure of importance.

\subsection{Empirical Sufficiency for Adaptive Layer Sampling}
While MGN is not a complete or causal measure of layer importance, our experiments demonstrate that it is sufficiently informative for guiding adaptive layer sampling. Across multiple models and tasks, we observe that:
(i) sampling layers according to MGN often outperforms uniform sampling baseline,
(ii) static importance estimates are inferior to dynamically updated MGN statistics, and
(iii) MGN exhibits task-dependent and training-stage-dependent patterns across different architectures.

These observations indicate that MGN captures meaningful optimization signals that are directly relevant to the layer selection problem addressed by GRASS. Therefore, MGN serves as an effective and efficient proxy for allocating update resources under memory-constrained fine-tuning settings.

\section{Additional Experiments}
\label{sec:additional_experiments}

\subsection{Layer-wise MGN on Different Models}
\label{sec:layer-wise_mgn}
\begin{figure*}[t]
  \includegraphics[width=0.495\linewidth]{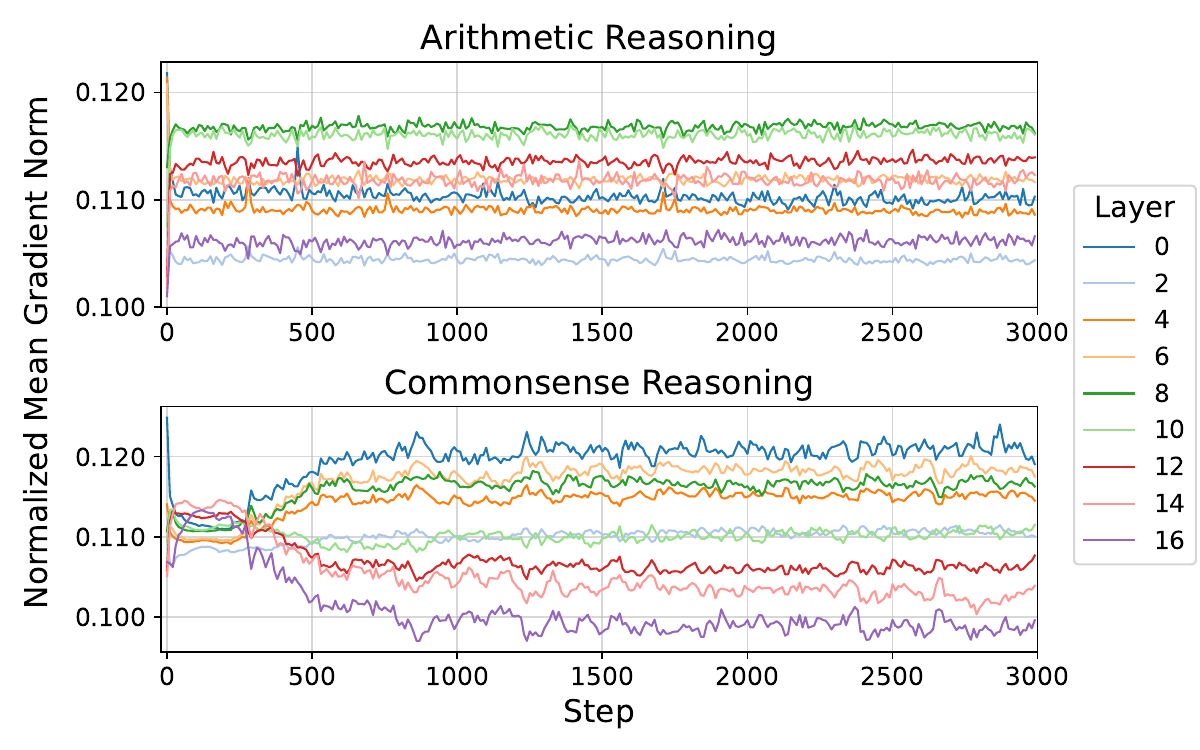} \hfill
  \includegraphics[width=0.495\linewidth]{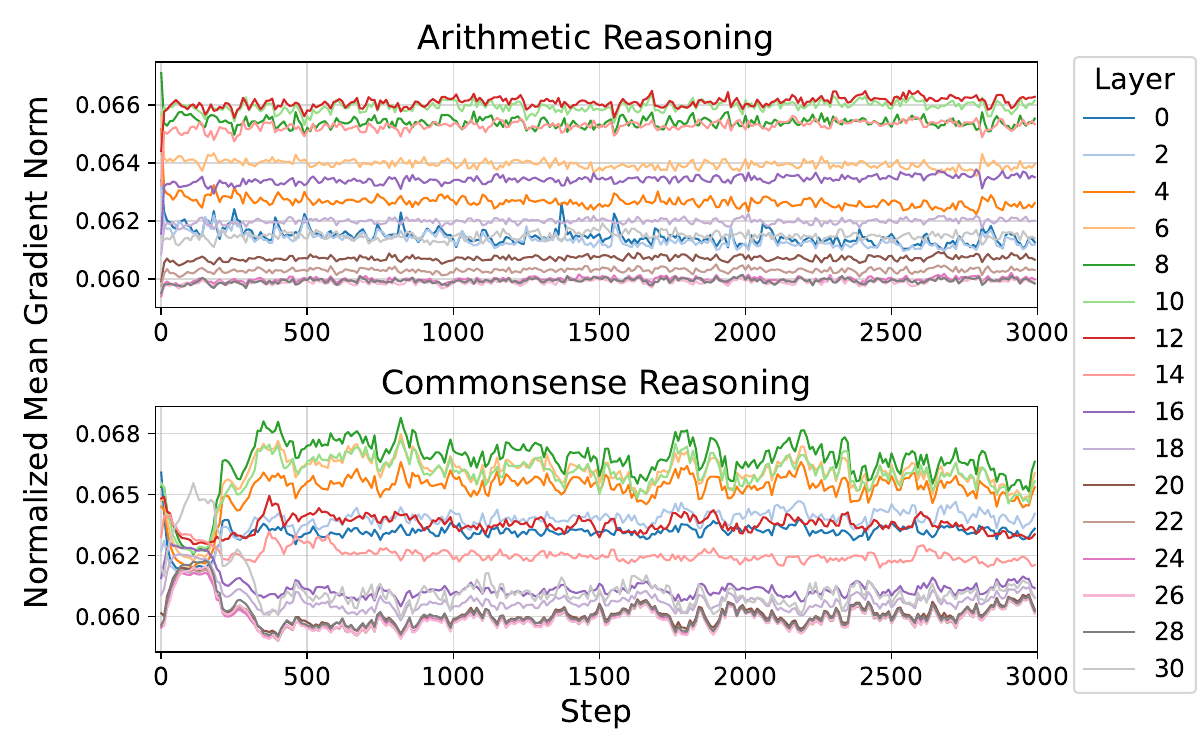}
  \caption {Comparison of layer-wise mean gradient norm across different datasets on Gemma-2B (Left) and LLaMA2-7B (Right).}
  \label{fig:additional_mgn}
\end{figure*}

We conduct further experiments on Gemma-2B and LLaMA2-7B to demonstrate that MGN is a task-aware indicator that captures varying layer-wise importance across downstream tasks and training stages. The results are shown in Figure~\ref{fig:additional_mgn}.

Similar to the observations on TinyLlama, we find that MGN values exhibit layer-wise differentiation and task-specific patterns across both models. In particular, the relative ordering of layer importance varies noticeably between arithmetic reasoning and commonsense reasoning tasks, indicating that different layers dominate the optimization dynamics depending on task characteristics. These results further support the validity of MGN as a indicator of layer importance, motivating its use for adaptive layer sampling in GRASS.

\subsection{Convergence}
\begin{figure*}[t]
  \includegraphics[width=0.329\linewidth]{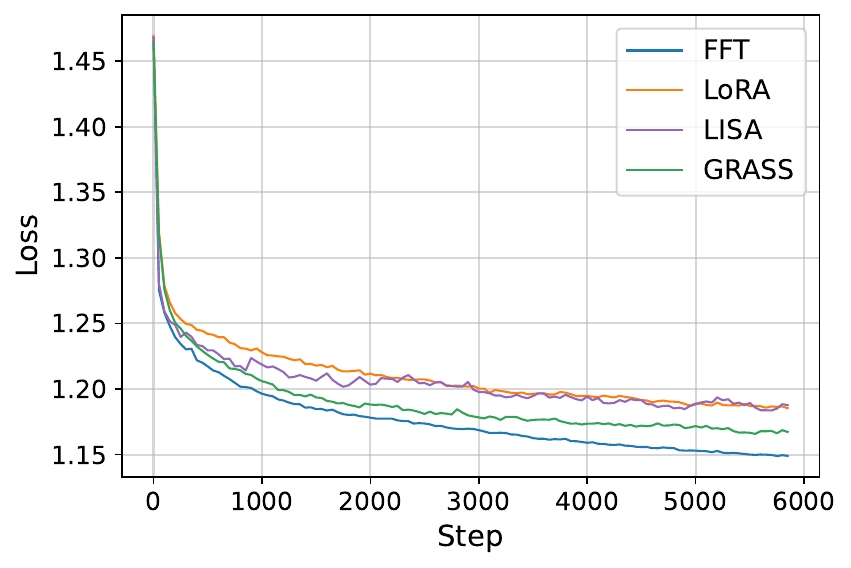} \hfill
  \includegraphics[width=0.329\linewidth]{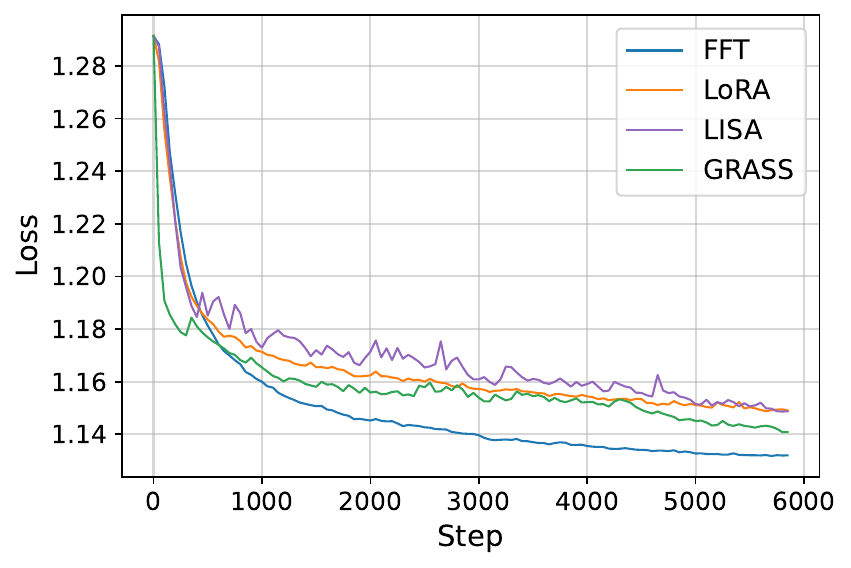} \hfill
  \includegraphics[width=0.329\linewidth]{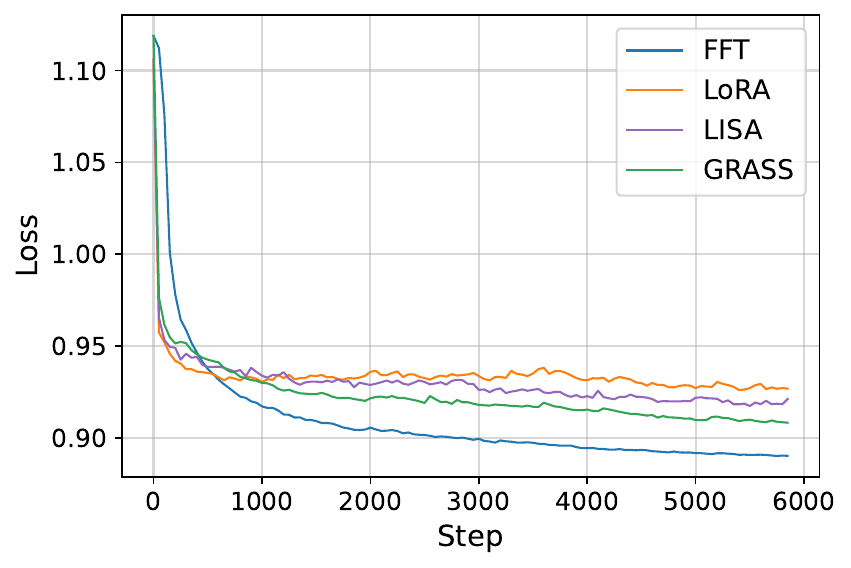}
  \caption {Loss curves of fine-tuning TinyLlama (Left), Gemma-2B (Middle), and LLaMA2-7B (Right) with different methods on Alpaca-GPT4 dataset.}
  \label{fig:loss_curve}
\end{figure*}

Figure~\ref{fig:loss_curve} illustrates the validation loss curves of different fine-tuning methods on the Alpaca-GPT4 dataset for TinyLlama, Gemma-2B, and LLaMA2-7B. Across all model scales, GRASS exhibits stable convergence behavior throughout training and consistently achieves lower validation loss in later training stages. The results indicate that GRASS effectively concentrates updates on the most influential components, leading to reliable convergence across different model sizes.

\subsection{Sensitivity to Probing Phase Length}
\label{sec:sensitivity_to_probing_phase_length}

\begin{table}[t]
\centering
\resizebox{0.8\linewidth}{!}{%
\begin{tabular}{@{}c|ccccc@{}}
\toprule
$T_p$ & 25 & 50 & 100 & 150 & 200 \\ \midrule
Avg. & 58.95 & 59.31 & 59.33 & \textbf{59.59} & 58.20 \\ \bottomrule
\end{tabular}%
}
\caption{Average accuracy on arithmetic reasoning tasks with varying lengths of probing phase.}
\label{tab:probing_phase}
\end{table}

At the start of training, GRASS employs a short probing phase of length $T_p$ to initialize layer-wise MGN statistics. To examine the sensitivity of GRASS to the choice of $T_p$, we vary $T_p$ from 25 to 200 steps and report the average accuracy on arithmetic reasoning benchmarks.
As shown in Table~\ref{tab:probing_phase}, increasing $T_p$ improves performance up to a moderate range, peaking at $T_p=150$. When the probing phase is too short (e.g., $T_p=25$), the initial MGN estimates can be noisy due to insufficient gradient observations, leading to slightly degraded performance. On the other hand, overly long probing phases (e.g., $T_p=200$) may delay the onset of adaptive layer sampling and reduce effective training time for optimization, resulting in mild performance degradation.
However, the performance variance between $T_p=50$ and $T_p=150$ remains small, indicating that GRASS is not sensitive to the choice of $T_p$ as long as it falls within a reasonable range. This robustness allows $T_p$ to be set flexibly without careful tuning, and we use $T_p=150$ as the default setting in all experiments.

\end{document}